\documentclass[11pt]{article}
\usepackage{authblk}

\newlength{\defbaselineskip}
\usepackage{graphicx}
\usepackage{amsmath}
\usepackage{amssymb}
\usepackage[caption=false,font=footnotesize]{subfig}

\DeclareMathOperator*{\argmax}{arg\,max}

\usepackage{algorithmic}
\usepackage{algorithm}

\begin{document}

\title{
The Effectiveness of Variational Autoencoders for Active Learning 
}

\author{Farhad Pourkamali-Anaraki\\Department of Computer Science, University of Massachusetts Lowell, MA, USA \and Michael B. Wakin\\Department of Electrical Engineering, Colorado School of Mines, CO, USA}


\maketitle

\begin{abstract}
	The high cost of  acquiring labels is one of the main challenges in deploying supervised machine learning algorithms. Active learning is a promising approach to control the learning process and address the difficulties of data labeling by selecting labeled training examples from a large pool of unlabeled instances. In this paper, we propose a new data-driven approach to active learning by choosing a small set of labeled data points that are both informative and representative. To this end, we present an efficient  geometric technique to select a diverse core-set in a low-dimensional latent space obtained by training a Variational Autoencoder (VAE).  Our experiments demonstrate an improvement in accuracy over two related techniques and, more importantly, signify the representation power of generative modeling for developing new active learning methods in high-dimensional data settings. 
\end{abstract}

\section{Introduction}
\label{sec:intro}

In many machine learning applications such as medical diagnosis, a major challenge is to collect sufficient amounts of supervised training data because the labeling process tends to require domain expertise and immense amounts of computational/experimental resources.
A promising approach to overcome this problem is \textit{active learning} \cite{cohn1996active,settles2012active,gal2017deep,Krishnamurthy2019}, which focuses on practical ways to choose a small subset of the data for labeling to train  accurate predictive models. In the pool-based active learning setting, we have access to a large set of unlabeled instances, as data collection is often straightforward.  The goal is then to query a small number of labels from an oracle, e.g., a human annotator. 
A recent work defined active learning as a core-set selection problem to improve the performance of existing methods~\cite{sener2017active}. Core-set construction has been a promising technique for large-scale learning such as classification and clustering \cite{munteanu2018coresets,lucic2017training2,pourkamali2019large}. 

From a geometric perspective, the idea behind  core-sets in unsupervised settings is to find  diverse subsets  that best cover the entire data. Thus, in the context of active learning with a limited labeling budget, it is reasonable to query labels for only  representative examples to train supervised models.  However, most existing works,  including \cite{sener2017active}, attempt to find such set covers in the original input space.

A remarkable drawback of prior work  is that geometric methods often have poor performance when analyzing high-dimensional data sets, such as image data, because of the inefficiency of similarity measures such as the Euclidean norm \cite{min2018survey}. Moreover, covering methods are often NP-hard and require multiple initializations.  Therefore,  solving related geometric optimization problems in the high-dimensional input space is computationally expensive and impractical, specifically for a large number of unlabeled instances. 

In this paper, we propose a new approach to active learning by \textit{learning} a mapping from the observed space to a lower-dimensional latent space. To learn an efficient compression of the input data, we use Variational Autoencoders (VAEs) \cite{kingma2013auto,rezende2014stochastic,lopez2018information}. VAEs are popular techniques to find complex latent-variable generative models for  high-dimensional data sets such as image data. 
The success of VAEs stems from the representation power of neural networks for approximating complex functions and variational inference for Bayesian models \cite{zhang2018advances}. 

Our second contribution is to design an efficient and easy-to-implement method for finding core-sets in the latent space of a VAE. The proposed method uses K-means clustering \cite{pourkamali2017preconditioned} to recognize regions of interest with possibly homogeneous labels. 
Subsequently, the learner trains a supervised model, e.g., a classifier, in the latent space. Due to the generative nature of our framework, we can map data points outside the original pool into the latent space. Therefore, our active learning approach complements existing works focusing on choosing representative examples in the input space and paths the way for future improvements in terms of achieving higher accuracies and computational savings. 

We also present  experiments on the MNIST database of handwritten digits \cite{lecun1998gradient} to verify the performance of our proposed method empirically. The results of an active learning algorithm are typically depicted by a curve measuring the trade-off between the  number of labeled points and classification accuracy.  Unlike most existing works, we do not assume a fixed hypothesis or model for the classification task because an appropriate classifier is usually not known a priori. To have a  fair comparison in our experiments, instead, we allow our active learning method to decide the best classifier according to the labeling budget.   

The rest of this paper is organized as follows. Section \ref{sec:review} formulates the active learning problem
and provides a brief overview of prior art. Section  \ref{sec:method} introduces the proposed method. Extensive experiments in Section \ref{sec:exper}  demonstrate the effectiveness of VAEs for active learning, while Section \ref{sec:conclusion}
provides concluding remarks.

\section{Relation to Prior Work}
\label{sec:review}
 Suppose we are given a set of $n$ \textit{unlabeled}  points $\mathcal{X}=\{x_1,\ldots,x_n\}$ in $\mathbb{R}^D$ with a label space $\mathcal{Y}=\{1,\ldots,C\}$, corresponding to classification with $C$ classes. We then acquire labels for a small set of $B$ points from $\mathcal{X}$ to train a classifier. Thus, the labeling budget is $B<n$ in the pool-based active learning framework.
 
 In the classical setting, active learning methods choose a single  point from the pool of unlabeled instances $\mathcal{X}$ at each iteration until requesting labels for a total of $B$ points. To this end, a \textit{fixed} model actively selects the data points for which the most uncertain~\cite{lewis1994heterogeneous,tong2001support,joshi2009multi,gal2017deep,beluch2018power}. Intuitively, such a sampling strategy will choose more informative data points compared to sampling uniformly at random from the unlabeled pool $\mathcal{X}$ \cite{you2018scalable}. For example, a binary classifier, such as logistic regression, selects a point at each round whose posterior probability is closest to $0.5$, i.e., has the maximum entropy.  

Active learning methods using uncertainty information have two main shortcomings. For large-scale data, it is impractical to query a single point from $\mathcal{X}$ at each active learning iteration due to the high cost of sampling, e.g., calculating the entropy of the entire unlabeled set. Also, the model has to be retrained after acquiring each label, which can lead to computational and statistical inefficiencies. For example, a single point is likely to have no statistically significant impact on the accuracy of the trained classifier. The second issue is that highly correlated queries across different iterations of active learning reduce the overall efficiency, as a large part of the budget may focus on repeatedly choosing nearby points \cite{yang2015multi,pinsler2019bayesian}. This problem is exacerbated in  multi-class classification problems as standard active learning methods are often biased towards certain regions and miss critical information about the distribution of the unlabeled pool $\mathcal{X}$. 

Therefore, there is consensus now on the need for improved active learning methods that take into account the \textit{diversity} of selected points for labeling. The authors of \cite{sener2017active}  proposed a geometric approach, which finds a small subset  that best covers the entire unlabeled pool $\mathcal{X}$. To be formal, given a candidate set $\mathcal{S}$ with size less than $B$, the next point to query is chosen as follows:
\begin{equation}
\argmax_{x_i\in\mathcal{X}\setminus\mathcal{S}} \min_{x_j\in\mathcal{S}} \Delta(x_i,x_j),
\end{equation}
where $\Delta(\cdot,\cdot)$ is a distance metric in the input space $\mathbb{R}^D$. A critical component of this geometric approach is the choice of $\Delta$. Previous work considered the Euclidean distance for simplicity. However, it is still an open question whether we can find more meaningful metrics for complex high-dimensional data. 

In this paper, we propose a new approach to tackle this problem based on  learning an informative and low-dimensional data representation using the unlabeled data set $\mathcal{X}$ before acquiring labels for $B$ points. Hence, the proposed framework has several advantages: it will lead to significant improvements in terms of computational and statistical  efficiencies, and furthermore, it can be seamlessly integrated into other frameworks focusing on finding diverse subsets of data points for active learning or other purposes. 

\section{Proposed Method}
\label{sec:method}
In this section, we explain why variational autoencoders (VAEs) are effective tools to find a  small set of diverse examples from the unlabeled data set $\mathcal{X}$ in the pool-based active learning framework. To this end, we briefly review the main underlying ideas behind VAEs. Next, we discuss our proposed method in the latent space provided by training a VAE on $\mathcal{X}$. While we focus on using the entire unlabeled set in this paper, an interesting future research direction is to train VAEs using a portion of unlabeled instances $\mathcal{X}$. 

VAEs are powerful generative models capable of learning unsupervised latent representations of complex high-dimensional data. In the VAE framework, one approximates the intractable posterior distribution over a set of latent variables, i.e., $p(z|x)$, with another distribution $q(z|x)$. The  Kullback-Leibler divergence between these two  distributions \cite{dai2018connections}, i.e., $D_{KL}(q(z|x)||p(z|x))$, is  minimized by maximizing a lower bound on the marginal log-likelihood  over the data in the following form when $p(z)$ is a predefined distribution such as an isotropic Gaussian:
\begin{equation}
\sum_{x\in\mathcal{X}} \underbrace{\mathbb{E}_{q(z|x)}[\log p(x|z)]}_{\text{reconstruction term -- decoder}} - \underbrace{D_{KL} (q(z|x)|| p(z))}_{\text{regularization term -- encoder}}.
\end{equation}
From an information theoretic perspective, the variables $z\in\mathbb{R}^d$, $d<D$, are latent representations of the input data points $x\in\mathcal{X}$. From the neural network viewpoint, VAEs consist of an encoder $q(z|x)$, a decoder $p(x|z)$, and a loss function which can be optimized by stochastic gradient descent. Therefore, the encoder network $q(z|x)$ takes as input data in $\mathbb{R}^D$ and maps it into a lower-dimensional latent representation in $\mathbb{R}^d$. This feature transformation process forms the main building block for our proposed active learning method by facilitating the design of effective geometric methods. 

The proposed geometric approach in the latent space of a VAE is an efficient method capable of providing a diverse set of $B$ points that cover the entire data. To be formal, we propose to employ a clustering algorithm, such as K-means clustering, to partition the latent representations $z_1,\ldots,z_n$ into $K$ clusters ($K<B$); without of loss of generality, we assume that $m=B/K$ is an integer. This step allows us to capture the underlying structure in the latent space; see  \cite{meila2018tell} for a theoretical discussion. Next, we sample $m$ points uniformly at random from each cluster to create a set of $B$  points for labeling. Hence, our approach approximately solves the following optimization problem  over a core-set $\mathcal{S}$ to cover the full data $\mathcal{Z}=\{z_1,\ldots,z_n\}$ in $\mathbb{R}^d$:
\begin{equation}
\min_{\mathcal{S}: |\mathcal{S}|\leq B}\max_{z_i\in\mathcal{Z}\setminus\mathcal{S}}\min_{z_j\in\mathcal{S}} \|z_i-z_j\|_2. 
\end{equation}
That is, our proposed method aims to  find $B$ data points in the latent space $\mathbb{R}^d$ such that the largest Euclidean distance between every point in $\mathcal{Z}$ and its nearest representative from the core-set $\mathcal{S}$ is minimized. As a result, learning a latent representation of the input data has two benefits. It leads to significant computational savings for high-dimensional data and considerable improvements in performance when finding a meaningful metric for measuring distances in the data space is a difficult task. Moreover, the introduced active learning method in this paper offers an elegant framework for massive data sets with thousands or even millions of points by employing non-uniform sampling methods \textit{within} each of the $K$ clusters (instead of the uniform sampling currently used). Specifically, the proposed framework allows us to incorporate uncertainty information concerning a model within each of the $K$ partitions to achieve further improvements. However, in this work, our main focus is to design an agnostic active learning algorithm that works for any hypothesis class \cite{dasgupta2008general}. 

\section{Experimental Results}
\label{sec:exper}
We demonstrate the effectiveness of VAEs for reducing the number of labeled data points that are required to get a specified  accuracy on the MNIST data set (loaded from Keras). To this end, we first present a visualization of the latent space corresponding to three classes and show the advantages of our proposed framework compared to existing techniques. The second experiment consists of four and five classes from the MNIST data set to further show that our framework is appropriate for multi-class classification problems. 

\textbf{Classification with three classes:}
 In this experiment, we consider the problem of classifying three digits, i.e., 0, 3, and 9, using data which contains $18,\!003$ training images and $2,\!999$ test images of size $28\times28$ ($D=784$). Test images are used only for reporting the classification accuracy and not for training VAEs or classifiers.  

\begin{figure}[t]

\begin{minipage}[b]{1.0\linewidth}
  \centering
  \centerline{\includegraphics[width=8cm]{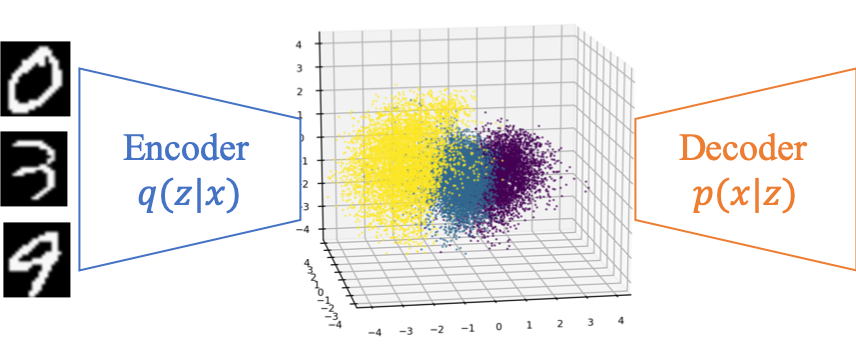}} 
  \centerline{(a) Visualizing  latent space obtained via VAE}\medskip 
\end{minipage}
\begin{minipage}[b]{1.0\linewidth}
  \centering
  \centerline{\includegraphics[width=8cm]{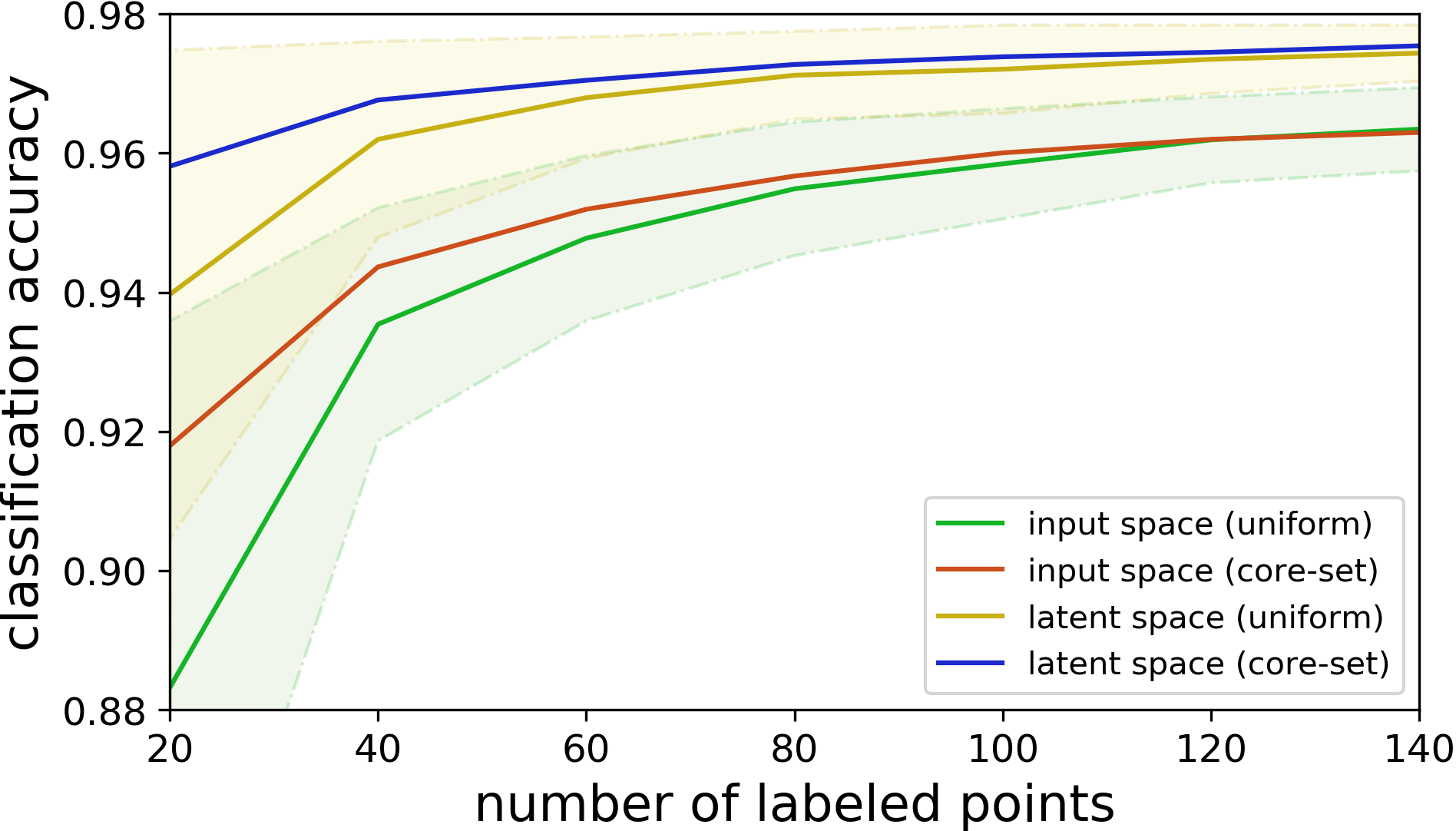}}
  \centerline{(b) Uniform sampling in input space vs.~latent space}\medskip
\end{minipage}
\begin{minipage}[b]{1.0\linewidth}
  \centering
  \centerline{\includegraphics[width=8cm]{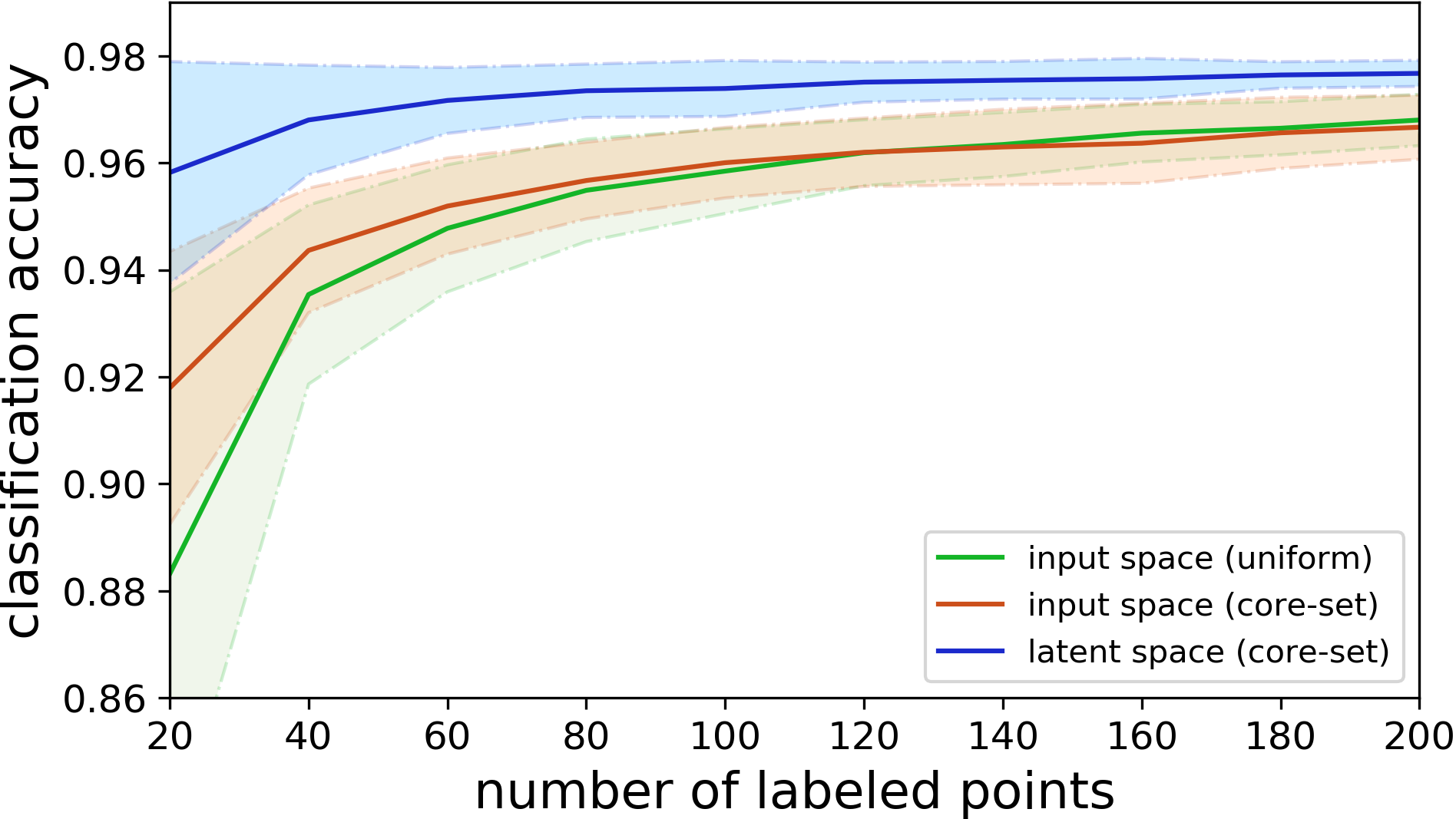}}
  \centerline{(c) Geometric sampling in input space vs.~latent space}
\end{minipage}
\caption{Active learning with three classes from MNIST.}
\label{fig:exper1}
\end{figure}

We use the unlabeled training data consisting of $18,\!003$ images to learn a latent space in $\mathbb{R}^d$ with $d=3$. To train a VAE for the image data, the encoder network consists of four convolutional layers \cite{papyan2017convolutional}, where both the height and width of convolution windows are set to be $3$. The numbers of filters in these four layers are $32$, $64$, $64$, and $64$, respectively. We use ReLU activation functions and the standard optimizer RMSProp. As a result, we map the $784$-dimensional input space into $\mathbb{R}^3$. Before reporting active learning results, we plot the latent space for the training data in Fig.~\ref{fig:exper1}(a), where we  color-code the embedding for visualization purposes (although we did not use those labels for training). As we see, the learned mapping provides an informative representation of the  data according to their known digit class.  

To exhibit the efficacy of VAEs in this example, we also perform K-means clustering in both input space and latent space when the number of clusters is $3$. While our main goal in this paper is to train supervised models, the clustering accuracy is a sensible measure to confirm that the obtained latent representation is indeed able to extract the underlying structure of the data. Normalized mutual information (NMI) \cite{pourkamali2019improved} is a popular clustering quality metric which ranges from $0$ to $1$, and larger values of NMI indicate the higher quality of clustering. In this experiment, the values of NMI in the input space and the latent space are $0.71$ and $0.88$, respectively. 

In the active learning setting, we do not have access to all $18,\!003$ labels. Hence, we choose varying number of labeled points $B$  from $20$ to $200$, and we then train a support vector classifier (SVC) on the labeled data (SVC loaded from scikit-learn, and supports multi-class classification). The classifier is optimized by cross-validation using GridSearchCV, and chooses between radial basis  and linear kernel functions. 

In Fig.~\ref{fig:exper1}(b), we report the mean and standard deviation of accuracy over $500$ independent trials for the number of labeled points ranging from $20$ to $140$. Here, we drop the standard deviation of geometric techniques to demonstrate the improvement achieved by just using a better data representation combined with uniform sampling. We observe that uniform sampling in the latent space consistently outperforms the same sampling method in the  input space. 
For example, the averaged classification accuracy in the latent space reaches $0.96$ with even $40$ labeled points, while at least $120$ labeled instances are required to reach the same accuracy without transforming the input data. Furthermore, uniform sampling in the latent space results in greater reduction in the standard deviation of classification accuracy. Thus, the low-dimensional representation achieved by the VAE is beneficial for reducing the number of labeled points to achieve a desired accuracy.  

In Fig.~\ref{fig:exper1}(c), we report the mean and standard deviation of core-set sampling methods in both input space and latent space with $K=20$ (number of partitions explained in Section \ref{sec:method}) and up to $200$ labeled points. We see that the mean accuracy of our proposed sampling technique in the latent space is about $0.96$ when $20$ labeled points are available. Even for larger values of labeled points, such as $200$ labeled points, our approach outperforms other related techniques such that the mean accuracy is $0.976$ with the standard deviation $0.003$.   

\textbf{Classification with four and five classes:}
In this experiment, we consider two problems of classifying four digits (0, 3, 7, and 9) and five  digits (0, 2, 3, 7, and 9) from the MNIST data set. We use the same network architecture as the previous case. However, we increase the dimension of latent space  slightly; $d=5$ for the data set with four digits and $d=7$ when we have five digits. Empirically, we observe that growing the latent space dimensionality leads to higher quality representations as the number of classes increases. 

\begin{figure}[t]

\begin{minipage}[b]{1.0\linewidth}
  \centering
  \centerline{\includegraphics[width=8cm]{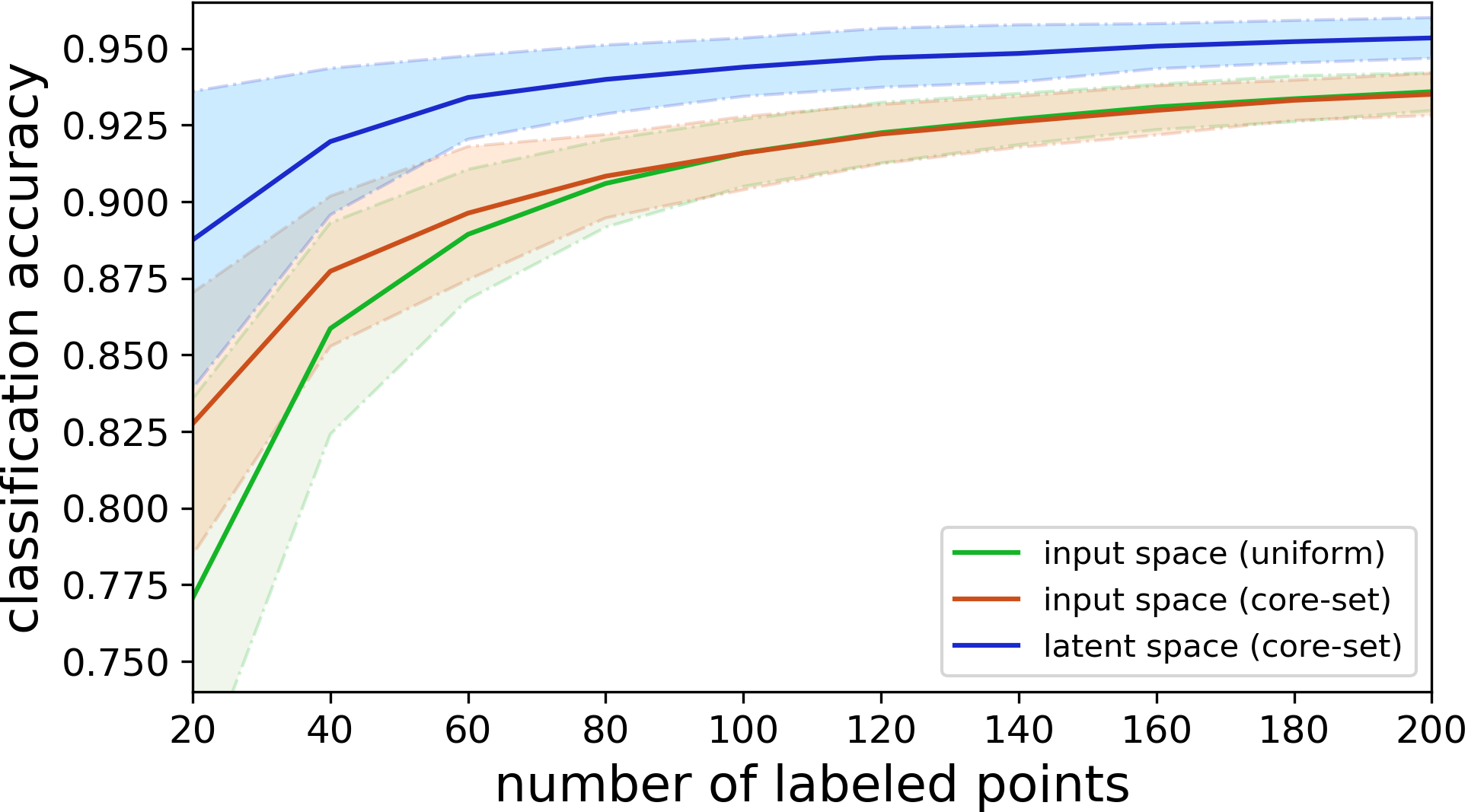}}
  \centerline{(a) Active learning with four classes}\medskip
\end{minipage}
\begin{minipage}[b]{1.0\linewidth}
  \centering
  \centerline{\includegraphics[width=8cm]{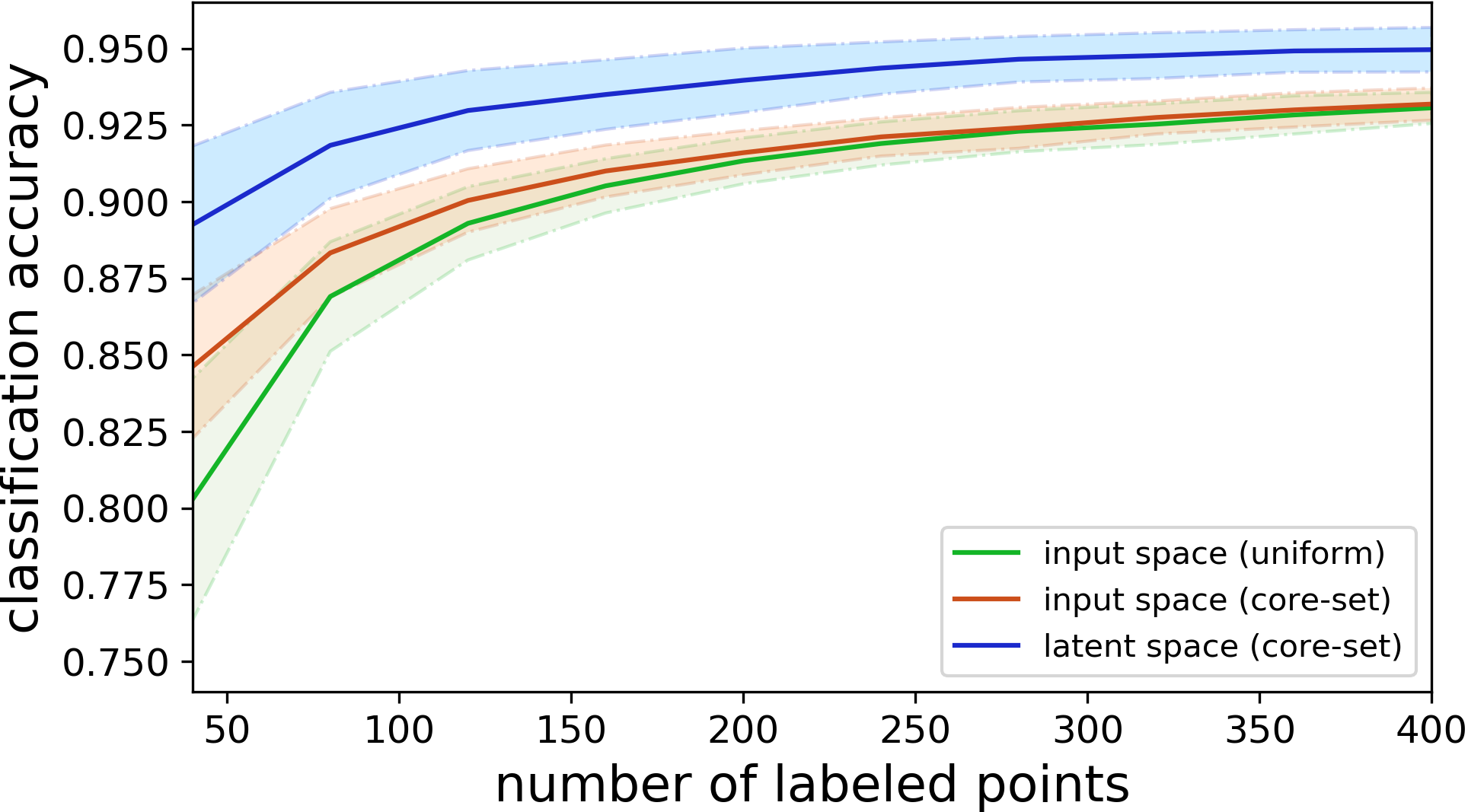}}
  \centerline{(b) Active learning with five classes}
\end{minipage}
\caption{Active learning with four and five classes.}
\label{fig:exper2}
\end{figure}

In Fig.~\ref{fig:exper2}, the mean and standard deviation of classification accuracy over $500$ independent trials are reported for varying values of $B$. In Fig.~\ref{fig:exper2}(a) and (b), we set $K=20$ and $K=40$, respectively. The main reason we choose $K=40$ for the case with five classes is to make sure that we have a few points from each class to train a classifier and tune corresponding parameters using cross-validation. Similar to the previous case,  our proposed sampling method in the latent space consistently outperforms other methods in the input space. Therefore, the proposed active learning method offers great potential to reduce the labeling cost to reach a certain accuracy.   

\section{Conclusion}
\label{sec:conclusion}
In this paper, we presented a new approach to active learning through finding compact representations of high-dimensional data using variational autoencoders. The proposed framework allows us to design efficient sampling strategies to query labels, which will result in developing cost-effective and accurate supervised models.

\bibliographystyle{plain}
\bibliography{refs}

\begin{thebibliography}{10}

\bibitem{beluch2018power}
W.~Beluch, T.~Genewein, A.~N{\"u}rnberger, and J.~K{\"o}hler.
\newblock The power of ensembles for active learning in image classification.
\newblock In {\em IEEE Conference on Computer Vision and Pattern Recognition},
  pages 9368--9377, 2018.

\bibitem{cohn1996active}
D.~Cohn, Z.~Ghahramani, and M.~Jordan.
\newblock Active learning with statistical models.
\newblock {\em Journal of artificial intelligence research}, 4:129--145, 1996.

\bibitem{dai2018connections}
B.~Dai, Y.~Wang, J.~Aston, G.~Hua, and D.~Wipf.
\newblock Connections with robust {PCA} and the role of emergent sparsity in
  variational autoencoder models.
\newblock {\em Journal of Machine Learning Research}, 19(1):1573--1614, 2018.

\bibitem{dasgupta2008general}
S.~Dasgupta, D.~Hsu, and C.~Monteleoni.
\newblock A general agnostic active learning algorithm.
\newblock In {\em Advances in Neural Information Processing Systems}, pages
  353--360, 2008.

\bibitem{gal2017deep}
Y.~Gal, R.~Islam, and Z.~Ghahramani.
\newblock Deep bayesian active learning with image data.
\newblock In {\em International Conference on Machine Learning}, pages
  1183--1192, 2017.

\bibitem{joshi2009multi}
A.~Joshi, F.~Porikli, and N.~Papanikolopoulos.
\newblock Multi-class active learning for image classification.
\newblock In {\em IEEE Conference on Computer Vision and Pattern Recognition},
  pages 2372--2379, 2009.

\bibitem{kingma2013auto}
D.~Kingma and M.~Welling.
\newblock {Auto-Encoding Variational Bayes}.
\newblock In {\em International Conference on Learning Representations}, 2014.

\bibitem{Krishnamurthy2019}
A.~Krishnamurthy, A.~Agarwal, T.~Huang, H.~Daum{{\'e}}, and J.~Langford.
\newblock Active learning for cost-sensitive classification.
\newblock {\em Journal of Machine Learning Research}, 20(65):1--50, 2019.

\bibitem{lecun1998gradient}
Y.~LeCun, L.~Bottou, Y.~Bengio, and P.~Haffner.
\newblock Gradient-based learning applied to document recognition.
\newblock {\em Proceedings of the IEEE}, 86(11):2278--2324, 1998.

\bibitem{lewis1994heterogeneous}
D.~Lewis and J.~Catlett.
\newblock Heterogeneous uncertainty sampling for supervised learning.
\newblock In {\em Machine Learning Proceedings}, pages 148--156. Elsevier,
  1994.

\bibitem{lopez2018information}
R.~Lopez, J.~Regier, M.~Jordan, and N.~Yosef.
\newblock Information constraints on auto-encoding variational {Bayes}.
\newblock In {\em Advances in Neural Information Processing Systems}, pages
  6114--6125, 2018.

\bibitem{lucic2017training2}
M.~Lucic, M.~Faulkner, A.~Krause, and D.~Feldman.
\newblock Training {G}aussian mixture models at scale via coresets.
\newblock {\em Journal of Machine Learning Research}, 18(1):5885--5909, 2017.

\bibitem{meila2018tell}
M.~Meila.
\newblock How to tell when a clustering is (approximately) correct using convex
  relaxations.
\newblock In {\em Advances in Neural Information Processing Systems}, pages
  7407--7418, 2018.

\bibitem{min2018survey}
E.~Min, X.~Guo, Q.~Liu, G.~Zhang, J.~Cui, and J.~Long.
\newblock A survey of clustering with deep learning: From the perspective of
  network architecture.
\newblock {\em IEEE Access}, 6:39501--39514, 2018.

\bibitem{munteanu2018coresets}
A.~Munteanu, C.~Schwiegelshohn, C.~Sohler, and D.~Woodruff.
\newblock On coresets for logistic regression.
\newblock In {\em Advances in Neural Information Processing Systems}, pages
  6561--6570, 2018.

\bibitem{papyan2017convolutional}
V.~Papyan, Y.~Romano, and M.~Elad.
\newblock Convolutional neural networks analyzed via convolutional sparse
  coding.
\newblock {\em Journal of Machine Learning Research}, 18(1):2887--2938, 2017.

\bibitem{pinsler2019bayesian}
R.~Pinsler, J.~Gordon, E.~Nalisnick, and J.~Hern{\'a}ndez-Lobato.
\newblock Bayesian batch active learning as sparse subset approximation.
\newblock In {\em Advances in Neural Information Processing Systems}, 2019.

\bibitem{pourkamali2019large}
F.~Pourkamali-Anaraki.
\newblock Large-scale sparse subspace clustering using landmarks.
\newblock In {\em IEEE International Workshop on Machine Learning for Signal
  Processing}, 2019.

\bibitem{pourkamali2017preconditioned}
F.~Pourkamali-Anaraki and S.~Becker.
\newblock Preconditioned data sparsification for big data with applications to
  pca and k-means.
\newblock {\em IEEE Transactions on Information Theory}, 63(5):2954--2974,
  2017.

\bibitem{pourkamali2019improved}
F.~Pourkamali-Anaraki and S.~Becker.
\newblock Improved fixed-rank {N}ystr{\"o}m approximation via {QR}
  decomposition: Practical and theoretical aspects.
\newblock {\em Neurocomputing}, 363:261--272, 2019.

\bibitem{rezende2014stochastic}
D.~Rezende, S.~Mohamed, and D.~Wierstra.
\newblock Stochastic backpropagation and approximate inference in deep
  generative models.
\newblock In {\em International Conference on Machine Learning}, pages
  1278--1286, 2014.

\bibitem{sener2017active}
O.~Sener and S.~Savarese.
\newblock Active learning for convolutional neural networks: A core-set
  approach.
\newblock In {\em International Conference on Learning Representations}, 2018.

\bibitem{settles2012active}
B.~Settles.
\newblock Active learning.
\newblock {\em Synthesis Lectures on Artificial Intelligence and Machine
  Learning}, 6(1):1--114, 2012.

\bibitem{tong2001support}
S.~Tong and D.~Koller.
\newblock Support vector machine active learning with applications to text
  classification.
\newblock {\em Journal of Machine Learning Research}, 2:45--66, 2001.

\bibitem{yang2015multi}
Y.~Yang, Z.~Ma, F.~Nie, X.~Chang, and A.~Hauptmann.
\newblock Multi-class active learning by uncertainty sampling with diversity
  maximization.
\newblock {\em International Journal of Computer Vision}, 113(2):113--127,
  2015.

\bibitem{you2018scalable}
C.~You, C.~Li, D.~Robinson, and R.~Vidal.
\newblock Scalable exemplar-based subspace clustering on class-imbalanced data.
\newblock In {\em European Conference on Computer Vision}, pages 67--83, 2018.

\bibitem{zhang2018advances}
C.~Zhang, J.~Butepage, H.~Kjellstrom, and S.~Mandt.
\newblock Advances in variational inference.
\newblock {\em IEEE Transactions on Pattern Analysis and Machine Intelligence},
  pages 2008--2026, 2018.

\end{thebibliography}

\end{document}